\documentclass[conference]{IEEEtran}
\IEEEoverridecommandlockouts
\usepackage{cite}
\usepackage{amsmath,amssymb,amsfonts}
\usepackage{algorithmic}
\usepackage{graphicx}
\usepackage{textcomp}
\usepackage{xcolor}
\usepackage[pagebackref,breaklinks,colorlinks]{hyperref}
\def\BibTeX{{\rm B\kern-.05em{\sc i\kern-.025em b}\kern-.08em
    T\kern-.1667em\lower.7ex\hbox{E}\kern-.125emX}}
\begin{document}

\def\needtocite{\textcolor{red}{[CITATIONS]}}
\newcommand{\red}[1]{\textcolor{red}{#1}}

\title{Video-Based Autism Detection with Deep Learning
%{\footnotesize \textsuperscript{*}Note: Sub-titles are not captured in Xplore and should not be used}
%\thanks{Identify applicable funding agency here. If none, delete this.}
}

\author{\IEEEauthorblockN{Manuel Serna-Aguilera}
\IEEEauthorblockA{\textit{Dept. of EECS} \\
\textit{University of Arkansas}\\
Fayetteville, United States \\
mserna@uark.edu}
\and
\IEEEauthorblockN{Xuan Bac Nguyen}
\IEEEauthorblockA{\textit{Dept. of EECS} \\
\textit{University of Arkansas}\\
Fayetteville, United States \\
xnguyen@uark.edu}
\and
\IEEEauthorblockN{Asmita Singh}
\IEEEauthorblockA{\textit{Dept. of Food Science} \\
\textit{University of Arkansas}\\
Fayetteville, United States \\
as118@uark.edu}
\and
\IEEEauthorblockN{Lydia Rockers}
\IEEEauthorblockA{\textit{Dept. of Food Science} \\
\textit{University of Arkansas}\\
Fayetteville, United States \\
lrockers@uark.edu}
\and
\IEEEauthorblockN{Se-Woong Park}
\IEEEauthorblockA{\textit{Dept. of Kinesiology} \\
\textit{University of Texas at San Antonio}\\
San Antonio, United States \\
sewoong.park@utsa.edu}
\and
\IEEEauthorblockN{Leslie Neely}
\IEEEauthorblockA{\textit{Dept. of Educational Psychology} \\
\textit{University of Texas at San Antonio}\\
San Antonio, United States \\
Leslie.Neely@utsa.edu}
\and
\IEEEauthorblockN{Han-Seok Seo}
\IEEEauthorblockA{\textit{Dept. of Food Science} \\
\textit{University of Arkansas}\\
Fayetteville, United States \\
hanseok@uark.edu}
\and
\IEEEauthorblockN{Khoa Luu}
\IEEEauthorblockA{\textit{Dept. of EECS} \\
\textit{University of Arkansas}\\
Fayetteville, United States \\
khoaluu@uark.edu}
}

\maketitle

% Notes:
% IEEE Greentech is not double-blind review, so we can keep the author names visible in our submission

%****************************************
% Abstract
%****************************************
\begin{abstract}
%Autism Spectrum Disorder (ASD) can often make life difficult for children, therefore early diagnosis is necessary for proper treatment and care.
%Thus, in this work, we consider the problem of detecting or classifying ASD in children to aid medical professionals in early detection.
%To this end, we develop a deep learning model that analyzes video clips of children reacting to sensory stimuli, with the intent on capturing key differences in reactions and behavior between ASD and non-ASD patients.
%Unlike many works in ASD classification, their data consist of MRI data, which requires expensive specialized MRI equipment, meanwhile our method need only rely on a powerful but relatively cheaper GPU, a decent computer setup, and a video camera for inference.
%Results on our data show that our model can generalize well and can understand key differences in the distinct movements of the patients.
%This is despite limited amounts of data for a deep learning problem, limited temporal information available to the model as input, and even when there is noise due to movement.
Individuals with Autism Spectrum Disorder (ASD) often experience challenges in health, communication, and sensory processing; therefore, early diagnosis is necessary for proper treatment and care. In this work, we consider the problem of detecting or classifying ASD children to aid medical professionals in early diagnosis. We develop a deep learning model that analyzes video clips of children reacting to sensory stimuli, with the intent of capturing key differences in reactions and behavior between ASD and non-ASD participants. Unlike many recent studies in ASD classification with MRI data, which require expensive specialized equipment, our method utilizes a powerful but relatively affordable GPU, a standard computer setup, and a video camera for inference. Results show that our model effectively generalizes and understands key differences in the distinct movements of the children. It is noteworthy that our model exhibits successful classification performance despite the limited amount of data for a deep learning problem and limited temporal information available for learning, even with the motion artifacts.
\end{abstract}


%****************************************
% Intro
%****************************************
\section{Introduction}
\label{sec:introduction}
% Intro to the problem & brief discussion of data and method
Autism Spectrum Disorder (ASD) is a broad set of conditions where people have difficulty communicating or exhibit abnormal behavior.
These conditions arrive in early childhood, and, for effective care, early ASD detection is important for the well-being of patients.
Accurate ASD diagnosis is a difficult task, however, deep learning can be leveraged to assist doctors with performing a more informed diagnosis.
We therefore introduce a deep-learning-based model that takes as input only videos of children reacting to different stimuli, and learn from the distinct reactions of ASD and neurotypical (NT) children to make accurate predictions without the need of specialized equipment such as MRI machines that cost hundreds of thousands of dollars. % citation needed?
Our training and testing data is purely video-based and acquired with a video camera, in contrast to several ASD detection works which rely on different varieties of MRI acquisition which is expensive, time consuming, and may not be immediately available to all communities.
The model consists of two convolutional neural network (CNN) backbones tasked with understanding ASD-related features and facial expression-related features, and this information is captured by a temporal transformer, connecting the spatial information across the frames in the temporal dimension.
To our knowledge, no other (public) video-based approaches like ours exist for ASD detection.

%****************************************
% Related Work
%****************************************
\section{Related Work}
\label{sec:related-work}

% TODO: trim this section first, and maybe intro to make further sections fit to 2 pgs

% Related works: image-, video-, etc.-based, mri-based
While deep learning has demonstrated significant success in various computer vision tasks \cite{truong2022direcformer, nguyen2023brainformer, nguyen2023fairness, he2016deep, swin-transformer-paper-2021, nguyen2023insect, nguyen2021clusformer, nguyen2023micron, nguyen2020self}, research in ASD detection or classification is somewhat scarce, and to our knowledge, no other works similar to ours exist.
%Jaby et al. \cite{image-asd-paper-2023} use RGB images as inputs, pass it through several different transformer-based \cite{attention-vaswani-paper-2017} sub-models \cite{vit-paper-2020, swin-transformer-paper-2021, mehta2022mobilevit} and perform classification via a support vector machine with their own image ASD dataset called FADC.
A closely-related work by Jaby et al. \cite{image-asd-paper-2023} use images with a Transformer-based model \cite{vit-paper-2020, swin-transformer-paper-2021, mehta2022mobilevit}, but do not make use of crucial temporal information for behavior analysis.
%Images by themselves, however, do not afford a model crucial temporal information that allows for understanding of ASD-related behaviors and actions.
%Washington et al. \cite{activity-recognition-asd-detection-2021} frame the problem of ASD detection under activity recognition, where certain actions belong more to ASD patients, and others not so.
%Their method is based on long short-term memory (LSTM) networks \cite{lstm-paper-1997} to make their prediction based on key movement.
Works such as Washington et al. \cite{activity-recognition-asd-detection-2021} use recurrent networks \cite{lstm-paper-1997} to perform classification based on particular behavior patterns--an activity recognition problem setting.
%This work contrasts with our method as our input data explicitly shows a stimulus evoking reactions in a more controlled environment and the face is meant to be more clearly visible, reducing the impact of noise.
%This work contrasts with our method as our input data explicitly shows a stimulus evoking reactions in a more controlled environment and the face is meant to be more clearly visible, reducing the impact of noise.
%Other research use eye gaze or eye movement to detect ASD, where problem is to find out eyes move and where subjects pay attention to in a scene.
Several works use eye gaze for ASD detection \cite{Jiang_2017_ICCV, fang-gaze-cnn-lstm-2020, eye-track-asd-detetion-mazumdar-2021, eye-tracking-DL-asd-detection-ahmed-2022} where they focus on tracking where a subject looks to in an image as an indicator for ASD under a classification problem.
Other work \cite{Chen_2019_ICCV} have ASD and NT subjects take the photos themselves and track the gaze and analyze photo-taking behaviors.
%Work by Chen et al. \cite{Chen_2019_ICCV} have ASD and NT subjects take the photos themselves and track the gaze, with the main idea of subject types' attention being significantly different for proper ASD classification.
% -----> Rewrite this: "This work contrasts with our method as our input data explicitly shows a stimulus evoking reactions in a more controlled environment and the face is meant to be more clearly visible, reducing the impact of noise."
By contrast, we analyze ASD-related behaviors by explicitly evoking reactions in more controlled settings. % keep?
Another class of methods analyze brain MRI data to find differences between ASD and NT patient brain activity.
%Common approaches for ASD classification use MRI data to find differences between ASD and NT patients in the brain.
%This class of works use different types of magnetic resonance imaging (MRI).
Some works use functional MRI (fMRI), which shows minute changes in blood flow in the brain, and deep learning-based approaches to extract important features as to which regions in the brain pertain more to ASD \cite{pyramid-kernel-asd-diagnosis-2023, hu-fmri-semi-supervised-lstm-asd-detection-2023, fmri-autoencoder-fscore-asd-detection-2023, fmri-asd-detection-cnn-2020, mri-asd-detection-rakic-2020}.
Other MRI works use resting state fMRI (rs-fMRI) and deep learning to classify patients \cite{deep-svm-meta-learning-asd-diagnosis-2023, convolution-kernel-asd-detection-2021}.

%****************************************
% Data
%****************************************
\section{Data Collection}

The dataset collection has been a collaboration between The University of Arkansas (UARK) and The University of Texas at San Antonio (UTSA).
The collective data consists of two distinct sets, the UARK split and UTSA split. %, for training and testing purposes, respectively.
To record our video data, the participants sit down and look at a screen, towards the camera, with a neutral face.
The stimulus item is then provided, the participant interacts with it, and then the camera captures the reactions--the movements of the subject, their head, and their facial expressions.
It is these important aspects of our data that the model aims to learn; it learns how the complex and subtle movements and expressions contribute to ASD classification.

% UARK data
The UARK data split was assembled by the Food Science Department at the University of Arkansas.
This dataset consists of several videos of 30 subjects reacting to different sensory stimuli.
Half of the entire group have ASD, and the other half are NT.
The taste and the smell senses were tested for this data split.
There are five taste stimuli, i.e., the subjects were given five items to taste: sucrose, caffeine, salt, citric acid, and quinine.
For the smell stimuli, there are eight samples: cabbage, peppermint, garlic, caramel, mushroom, citrus, vanilla, and fish. 
There are 150 taste experiment videos available to us, totaling approximately 98,000 frames, or 653 frames per video, on average.
There are 240 smell experiment videos available, totaling approximately 153,000 frames, or 637 frames per video, on average, available to us.
About halfway into each video, the interaction with the stimuli occur, where the seconds just after are the most important frames.

% SA data
The UTSA data collection is the same as the UARK data.
More senses are considered on top of smell and taste: auditory, texture, vision, and multimodal (of multiple senses) stimuli.
Note that we do not currently consider the ``extra'' senses at the moment.
This data split contains videos of 36 subjects, where 25 have ASD and 11 are NT.
There are 191 taste experiment videos available.
Meanwhile, there are 333 smell experiment videos available, total about 300,000 frames.
All videos in this dataset contain 900 frames, on average.

%****************************************
% Method
%****************************************
%
% Camera-ready version: I tried to clarify the data flow and structure of the model here instead of trying to explain important details in the experimental results section.
%
\section{Method}
\label{sec:method}
% Encoder part
%Our method is a deep learning model, with slices of video clips (i.e., a subset of the frames, not all the frames) as its input.
Our method is a deep learning model with slices of consecutive frames as its input.
Note that we sample the slices from a single video multiple times, so as to have more temporal context per video but not load the entire video, which in our experiments helps with generalization and efficiency.
%Given the frames, two specialized branches that extract global and local information, comprising of convolutional neural networks (CNNs), act as our backbones for feature extraction.
Given the video frames, two specialized backbone models, dubbed the main and facial expression (FER) models, extract two similar but distinct kinds of spatial features.
For main backbone, we crop the faces from the frames, and this maintains movement with sufficient action information, and the main backbone learns from this movement, not just facial structures.
%that we intend the model to learn from.
This allows for learning features related to movement in each frame that distinguish ASD and NT patients. 
In parallel, in the FER backbone, we perform face alignment on the input frames to allow the model to properly analyze the structure of the faces without noise from movement.
Thus, this model is solely focused on the expressive regions of the subjects' faces.
The output from these branches are movement and facial--spatial--features that are then fed to the decoder.

% Decoder part
The separate spatial features are concatenated together and then fed into our decoder--a temporal transformer \cite{vit-paper-2020} that learns how the frame information relate to each other and outputs the classification tokens. 
We then use a fully-connected network (or MLP) to output the probabilites for the NT and ASD classes. 
When we sample video slices, we average the probability predictions to obtain a final prediction.
%\red{The model processes data as follows.}
%\red{The features from both CNN backbones are concatenated along the feature dimension (1280 for our specific models).} 
%\red{This latent data is fed as input to the temporal transformer, which output the classification tokens.}
%\red{Finally, an MLP outputs the ASD class probabilities.}

%****************************************
% Evaluation
%****************************************
\section{Experimental Results}

%\begin{table}[htbp]
%\caption{Model Evaluation stimulus reaction video data}
%\begin{center}
%\begin{tabular}{|c|c|c|}
%\hline
%&\multicolumn{2}{|c|}{\textbf{Dataset}} \\
%\cline{2-3} 
%\textbf{Eval. Metric} & \textbf{UARK}& \textbf{SA} \\
%\hline
%Accuracy (\%) & 83.95 & x \\
%\hline
%F1 score & x & x \\
%\hline
%\multicolumn{3}{l}{}
%\end{tabular}
%\label{tab1}
%\end{center}
%\end{table}

% Implementation and hyperparams
We implement our framework using Pytorch \cite{pytorch-paper}.
We train and test our model with the UARK dataset using five $k$ folds, with four folds used as training samples, and the last fold for testing.
Our model trains with the AdamW optimizer \cite{adamw-optimizer-paper} with a cosine annealing scheduler \cite{cosine-scheduler-paper}, a learning rate 0.0001, a weight decay of initially 0.0001, and a minimum learning rate of 0.00001; the cross entropy loss is calculated with the MLP output and the true labels per video.
The batch size is set to 4 (8 also tends to do relatively well, but at the cost of less frames due to limited GPU memory) and we train the model for 40 epochs on one Quadro RTX 8000 GPU with 48 GB.

% Model details
In practive, we perform face detection, landmark detection (for face alignment), and pose angle estimation as a preprocessing step. 
The head pose angles (yaw, pitch, and roll) are computed as an additional preprocessing step to filter out frames where the face is not easily visible, and this helps our model consistently extract useful information.
%For the head pose angles to filter out cases where the face is not visible to help these networks.
All input images have the spatial dimensions of $224 \times 224$.
For the main ASD feature extraction branch, we use an EfficientNet B0 CNN \cite{efficientnet-paper-2020} that was pretrained on ImageNet \cite{russakovsky2015imagenet}, while the FER branch is a ResNet-18 \cite{resnet-paper-2015} that was pre-trained for the facial expression recognition task on MS-Celeb \cite{ms-celeb-dataset-paper-2016}.

%\red{
For evaluation, we compute the classification accuracy at the per-video level.
Given the probability of whether a video's subject has ASD, it is binarized into a yes (i.e., 1) or no (i.e., 0) label. %} %; this filters out many frames not useful for classification.}
%\red{To classify a video, we first remove frames with extreme head poses (in our case, head pose angles within $[-10, 10]$), and then we sample multiple slices per video. } % these details are already given later in the section
%\red{
For this work, we sample each video twice with 16 frames for each slice, totaling 32 frames (with more computational resources, the model can sample more frames, at the cost of GPU memory and processing time). %}
Thus, the test accuracy on the test dataset is 81.48\% with an F1 score of 0.7289.
The result indicates the model is able to generalize well to similar but unseen samples, despite limited data and the number of frames processed per video.
Sampling two times gives a good trade-off between speed, memory consumption, and performance.
Two slices of the video allow the model just enough context to make an accurate prediction and reduce the likelihood of the video slice giving noise or useless information.
We do note, however, that movement noise hampers performance, hence, we filter out frames where any head pose angle is out of the range $[-10, 10]$ to keep the data as controlled and noise-free as possible; this also affected the number of frames available from the UTSA set.
Thus, when we present the current model with video clips of a controlled setting, acheiving good performance.
Future work will involve addressing how to incorporate frames with more extreme head poses, allowing for the analysis of up to hundreds of more frames with critical information, and account for other kinds of noise like movement and occlusions of the face.

%****************************************
% References
%****************************************
\bibliography{bib/egbib}{}
\bibliographystyle{plain}

\end{document}